\documentclass[a4paper, 10pt, conference]{IEEEtran}
\IEEEoverridecommandlockouts
\usepackage{cite}
\usepackage{amsmath,amssymb,amsfonts}
\usepackage{algorithmic}
\usepackage{graphicx}
\usepackage{textcomp}
\usepackage{xcolor}
\def\BibTeX{{\rm B\kern-.05em{\sc i\kern-.025em b}\kern-.08em
    T\kern-.1667em\lower.7ex\hbox{E}\kern-.125emX}}
\begin{document}

\title{MeSLAM: Memory Efficient SLAM based on Neural Fields\\
}

\author{\IEEEauthorblockN{Evgenii Kruzhkov,
Alena Savinykh,
Pavel Karpyshev,
Mikhail Kurenkov, \\
Evgeny Yudin,
Andrei Potapov, and 
Dzmitry Tsetserukou}
\IEEEauthorblockA{\textit{ISR Laboratory, Skolkovo Institute of Science and Technology, Moscow, Russia}}
\IEEEauthorblockA{$\{$Evgeny.Kruzhkov, Alena.Savinykh, Pavel.Karpyshev, Mikhail.Kurenkov, \\
 Evgeny.Yudin, Andrei.Potapov, D.Tsetserukou$\}$@skoltech.ru}}

\maketitle

\begin{abstract}
Existing Simultaneous Localization and Mapping (SLAM) approaches are limited in their scalability due to growing map size in long-term robot operation. Moreover, processing such maps for localization and planning tasks leads to the increased computational resources required onboard.
To address the problem of memory consumption in long-term operation, we develop a novel real-time SLAM algorithm, MeSLAM, that is based on neural field implicit map representation. It combines the proposed global mapping strategy, including neural networks distribution and region tracking, with an external odometry system. As a result, the algorithm is able to efficiently train multiple networks representing different map regions and track poses accurately in large-scale environments. Experimental results show that the accuracy of the proposed approach is comparable to the state-of-the-art methods (on average, 6.6 cm on TUM RGB-D sequences) and outperforms the baseline, iMAP$^*$. Moreover, the proposed SLAM approach provides the most compact-sized maps without details distortion (1.9 MB to store 57 m$^3$) among the state-of-the-art SLAM approaches.

\end{abstract}


\section{Introduction}

\subsection{Motivation}

Nowadays, the area of autonomous robotics is developing at a very high pace. Its market constituted 1,61 billion USD in 2021 and is expected to grow 13 times by 2030 \cite{tmirob}. For the last decade, mobile robots have been successfully implemented in many areas including goods delivery \cite{protasov2021cnn}, warehouse logistic\cite{kalinov2019high, kalinov2020warevision, kalinov2021impedance, kalinov2021warevr}, autonomous transport, disinfection \cite{perminov2021ultrabot, mikhailovskiy2021ultrabot}, and agriculture \cite{karpyshev2021autonomous}. Autonomous robots are starting to work alongside humans executing tasks of increasing complexity. Such wide area of application requires robotic systems, including both hardware and software parts, to be robust, safe and efficient in challenging environments, e.g. in day and night conditions \cite{savinykh2022darkslam}. As a part of the software architecture, autonomous robots typically have a perception subsystem that mainly solves the problem of Simultaneous Localization and Mapping (SLAM). It generates a map and defines the position of the robot within the map at the same time. Thus, SLAM is a crucial task that must be solved accurately and efficiently to perform primary robot operation. Currently, many research collectives, both academic and industrial, are aimed at developing new SLAM approaches, designing task-driven methods and improving the existing pipelines to increase their robustness.


\begin{figure}[!t]
\centerline{\includegraphics[width=8.7 cm]{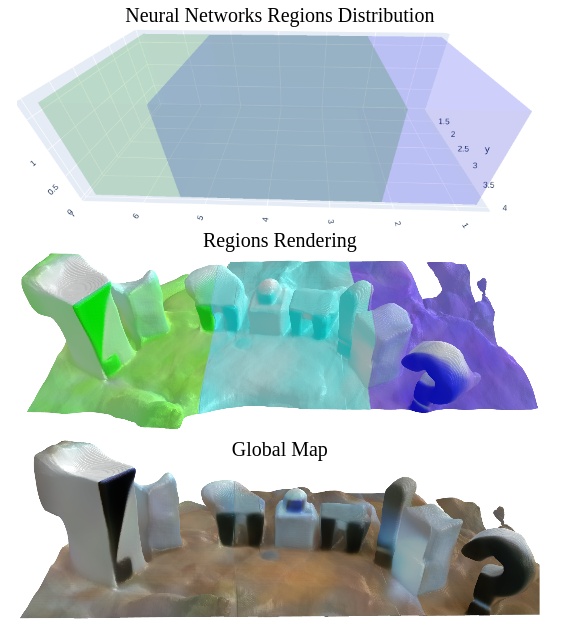}}
\caption{Distribution of regions assigned to a single neural network for rendering. Top: in green --- left region, in blue --- right region, in between --- regions intersection. In the middle: corresponding regions rendering. At the bottom: the final merged global map without visible stitching.}
\label{preview}
\end{figure}

\begin{figure*}[!t]
\centerline{\includegraphics[width=17.4 cm]{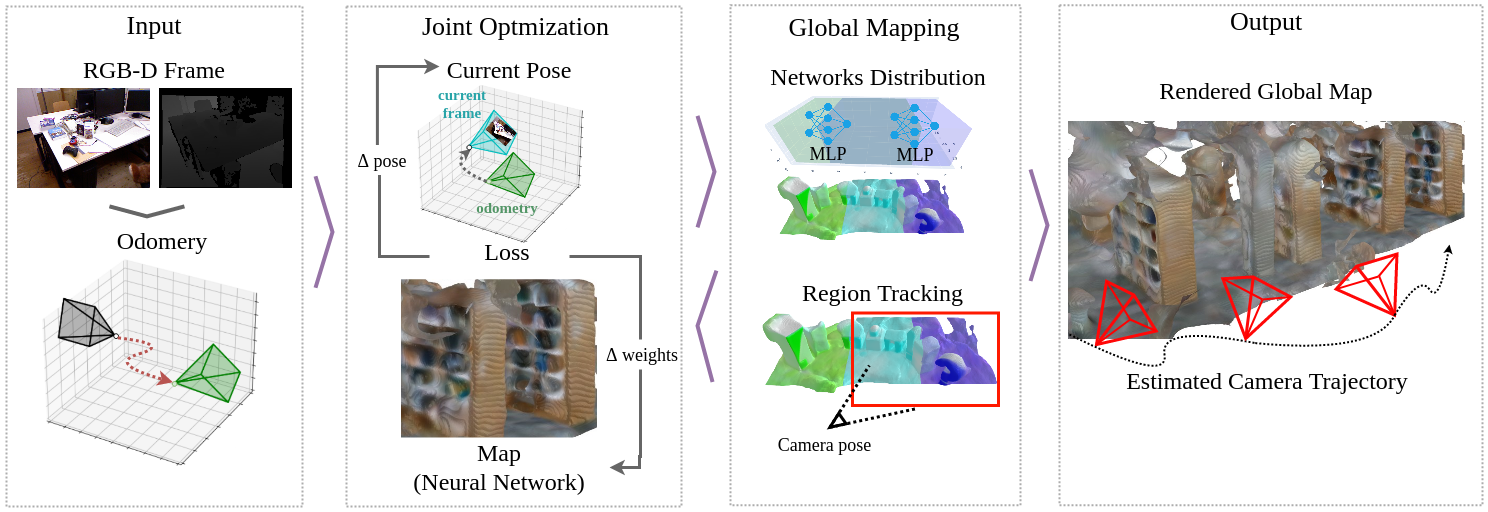}}
\caption{MeSLAM Overview.}
\label{sys_overview}
\end{figure*}

\subsection{Problem Statement}

One of the questions that arise in the SLAM problem is lifelong robot operation, which involves storage, modification, and access to large-sized maps, e.g. point clouds of cities. There are two major types of maps by structure: dense and sparse. Sparse maps contain an insignificant amount of data about the scene (objects on them are barely recognizable by humans); hence, they are light-weighted compared to dense maps. Contrariwise, dense maps include rich information about the environment; such maps are useful in static and dynamic object recognition, semantic segmentation and object tracking. Furthermore, they are understandable for the human perception that can facilitate the human-robot interaction area. However, memory consumption by dense maps is high, and their processing involves more computational resources to load and operate within. 
Thus, solving the SLAM problem implies a trade-off between required map density and available resources during both map creation and maintenance.

The high requirements for computational resources result in increased costs spent on data storage and other hardware, therefore enlarging the overall cost of robotics solution implementation. This work is aimed to research the question of achieving low memory consumption by dense mapping in SLAM systems for large-scale environments.



\subsection{Related Works}
\subsubsection{Sparse SLAM}
SLAM approaches that output sparse maps are regularly either visual-based or LIDAR-based. In general, the first group consists of four steps: feature-based frontend, local mapping that defines and tracks local keyframes, loop closure detection, and bundle adjustment (BA). State-of-the-art approaches like ORB-SLAM3 \cite{campos2021orb}, OV$^2$SLAM \cite{ferrera2021ov} and DXSLAM \cite{li2020dxslam} are designed in this paradigm. Although these methods are highly accurate, an empty featureless environment can easily distract their performance since they rely only on visual data. Moreover, the map consistency strongly depends on the number of features detected and matched, thereby, plain scenes result in very sparse maps or even loss of tracking.

Another group of SLAM methods that provide sparse maps is LiDAR-based solutions \cite{shan2018lego, wang2021f}. They lack the disadvantages inherent to visual-based solutions, yet the sparsity of the maps strongly depends on the LiDAR model used in a setup. The higher horizontal and vertical angles of view, and number of laser pairs in LiDAR result in denser maps.


\subsubsection{Dense SLAM}

As for dense SLAM approaches, KinectFusion \cite{newcombe2011kinectfusion} obtains a dense 3D map based on voxel representation by combining current depth images in the voxel space and using camera positions calculated by the ICP algorithm. Its improvement, Kintinuous \cite{whelan2012kintinuous}, provides dense mesh-based mapping of relatively large environments in real-time. However, these algorithms lack loop closure detection, which leads to higher accumulated errors and results in inaccurate 3D models. Another approach, ElasticFusion \cite{whelan2015elasticfusion}, provides dense globally consistent surfel-based maps of scalable environments without any post-processing steps, including pose graph optimization. Optimization, especially bundle adjustment, is rarely used or used with approximation in dense SLAM methods due to a huge number of variables. BAD SLAM \cite{schops2019bad} is a unique real-time dense SLAM approach that is able to implement bundle adjustment optimization for 3D map and positions without commonly used approximation. Although these dense methods have visually consistent 3D maps, they are unable to generate smooth continuous maps without holes. 

\subsubsection{Neural Field Approaches}
Both dense and sparse methods require higher memory consumption with the growth of map size or with the increased level of details obtained during the long operation on the same scene. Thus, the implicit scene representation based on neural fields gains large popularity due to the high capability of such methods to learn and store an environment in a compact form. The core idea of NeRF \cite{mildenhall2020nerf} is the optimization of neural radiance fields to render photorealistic novel views of cluttered scenes with high precision and continuous properties. However, the method is aimed at the reconstruction of small-sized objects (not rooms), and its main disadvantage is the significant time of training. Contrariwise, the kiloNeRF \cite{reiser2021kilonerf} approach has drastically decreased the time of NeRF reconstruction by exploiting multiple tiny Multi-Layer Perceptron (MLP) networks instead of the large one used in the NeRF method. This concept provides real-time operation without precision decrease, but the limited space of reconstruction is not addressed in the paper. Another approach, Block-NeRF \cite{tancik2022blocknerf} addresses the reconstruction of large-scale scenes. The algorithm is able to make the map of the city scene. However, both kiloNeRF and Block-NeRF methods solve the reconstruction task and not the SLAM one.

Based on NeRF ideas of neural radiance field optimization, iMAP \cite{sucar2021imap} proposes a real-time SLAM method that optimizes position simultaneously with the MLP training. The real-time operation is reached thanks to the reduction of the MLP size used in the original paper, and the use of depth map information. However, the utilization of a single MLP leads to the network ``forgetting" visited regions of the map, especially, in the case of large environments. NICE-SLAM \cite{zhu2021nice} introduces a SLAM method with hierarchical scene representation which is optimized along with pretrained geometric priors and results in a detailed reconstruction of a large indoor scene. Although decomposition of the environment results in a partial solution to the scaling problem, their discretization is tiny and, therefore, is hardly applicable for outdoors. Moreover, the usage of a pretrained network and training of the features instead results in a coarse reconstruction.


\subsection{Contribution}


We propose a novel SLAM algorithm for large-scale environment mapping with low memory consumption. It is achieved by the combination of recent achievements in neural implicit map representation with a proposed network's distribution strategy. 
The distribution module of our networks enables the usage of multiple MLP networks; each network's parameters implicitly represent a map of the region (local map) assigned to it. Moreover, the proposed distribution strategy provides the stitching of local maps into a global one without deformation, distorted overlapping and displacement; the example of obtained global map is shown in Fig. \ref{preview}.
Further, we propose a novel idea of combining external odometry with neural field-based optimization for efficient pose tracking in intersecting regions, and smooth stitching of local maps. 



\section{Methodology}

The overview of the proposed SLAM system is depicted in Fig. \ref{sys_overview}. RGB-D frames that include colored images and depth maps are transferred on algorithm's input; odometry is calculated for each RGB-D frame. Further, the odometry is combined with the joint neural-based optimization proposed in iMAP. During the joint optimization, the RGB-D frame is used to simultaneously refine the pose and neural network weights, which represent the current map of the region. During operation, the global mapping module distributes networks in the environment utilizing refined poses and determines the current working region involved in joint optimization. Other regions, not involved in the optimization at the current step, are stored. The whole environment or its parts can be reconstructed using the corresponding region set.

\begin{figure}[!htbp]
\centerline{\includegraphics[width=8.7 cm]{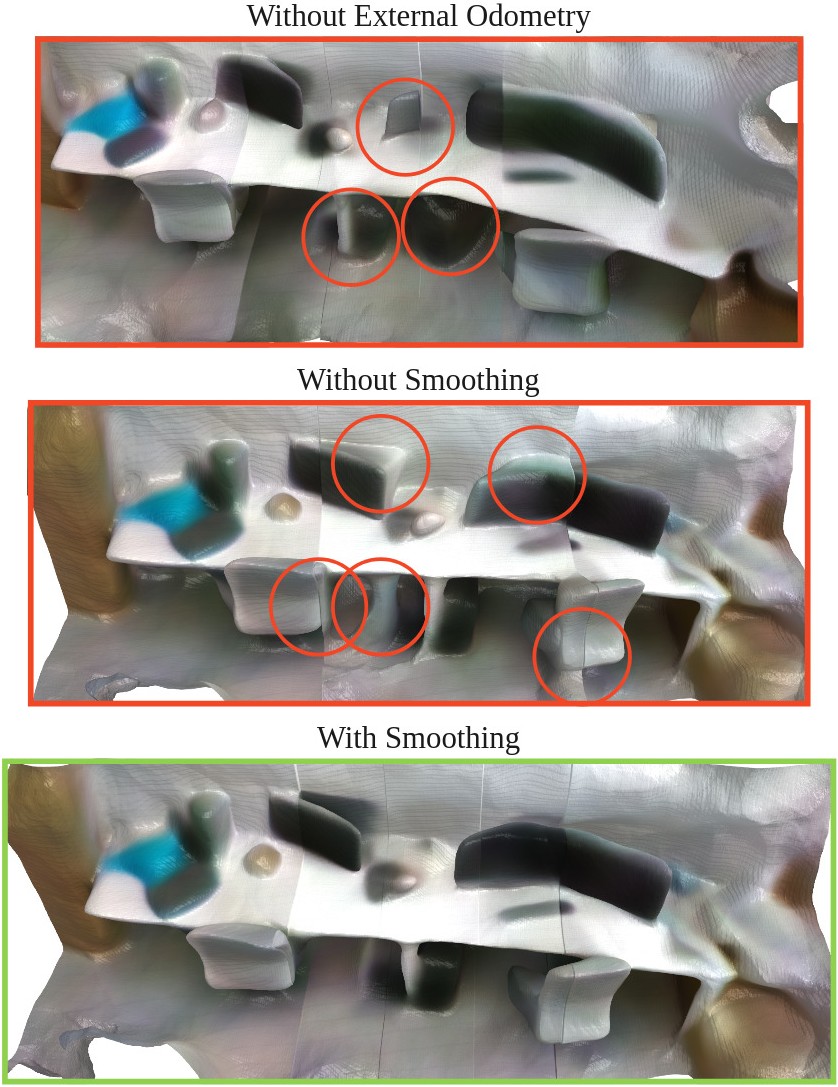}}
\caption{Influence of external odometry and smoothing on mapping. On the top: the smoothed mapping without external odometry --- displacements, false objects and artifacts. In the middle: non-smoothed mapping with external odometry --- many artifacts. At the bottom: mapping with external odometry and smoothing --- no displacements, no false objects, small artifacts.}
\label{reconstr-comp}
\end{figure}

\subsection{Odometry}

iMAP estimates the current pose using the previous one as initial and optimizes it according to the current frame. This approach strongly depends on the quality of neural network training achieved before the pose optimization begins; e.g. better performance in pose estimation can be reached if the network was pretrained for a while in a static start position. However, the use of external odometry decreases the relations between pose estimation accuracy and quality of network pretraining, and we use this property for robust tracking in intersecting regions. Fig. \ref{reconstr-comp} demonstrates that lack of odometry leads to imprecise global mapping with the presence of artifacts, displacements, and false objects.

The odometry for each RGB-D frame is calculated using Colored Point Cloud Registration \cite{park2017colored} algorithm. The method optimizes extended for RGB-D frame photometric and geometric objectives, providing point cloud alignment along both the normal and tangent directions. Then, the odometry calculated in each frame is used as the initial pose estimation before pose optimization to the current frame. 

\subsection{Joint Optimization}

\begin{figure}[!t]
\centerline{\includegraphics[width=8.7 cm]{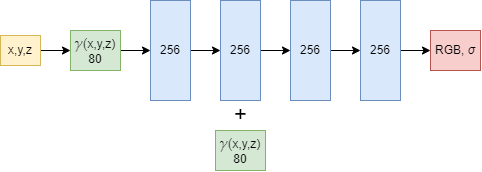}}
\caption{The architecture of MLP network. The yellow block is input data; the red one --- output. RGB data is obtained after sigmoid activation, while density $\sigma$ --- after ReLU activation. The green block is Gaussian positional embedding. Blue elements represent fully-connected MLP layers with ReLU activations. The numbers inside the blocks are dimensions. The symbol ``+" is concatenation.}
\label{network-arch}
\end{figure}

Joint optimization handles neural network parameters, representing the current map, and poses simultaneously. The map is refined by training Multi-Layer Perceptron (MLP) which parameters implicitly reflect the current map region. MLP architecture is represented in Fig. \ref{network-arch}.
The neural implicit representation is able to return color (RGB) and density ($\sigma$) for each point on the local map. To calculate them at each point of the local map, the network takes the point spatial coordinates $(x, y, z)$ as input (scaled in $[-1, 1]$ range), then applies Gaussian positional embedding $\gamma$, and, finally, returns the calculated values of color and density in the taken point after fully connected MLP body.

To refine the map, we use the NeRF strategy of rendering a predicted RGB-D frame from the current pose by integrating obtained colors and densities, and updating the MLP weights by minimizing geometric and photometric losses between the predicted frame and the real one. Further, the odometry pose at the current frame, instead of the previous pose in iMAP approach, is optimized by gradient descent together with MLP parameters. Thus, the current pose and current local maps are estimated by the joint optimization process.

The joint optimization process is inspired by the iMAP. However, we want to emphasize the importance of map coordinates $(x, y, z)$ scaling into $[-1, 1]$ range. This is achieved by utilizing the following transformation matrix:

\begin{align*}
  T_{wc} &= \left(\begin{array}{ccc|c}
        &       &      &      X_r - x \\
      \multicolumn{2}{c}{\smash{\raisebox{0.1\normalbaselineskip}{$R_{wc}$}}}
      &                &      Y_r - y \\
      &         &      &      Z_r - z \\
      \hline \\[-\normalbaselineskip]
     0  &  0 &   0    &  1
    \end{array}\right),
\end{align*}

where $R_{wc}$ is the rotation between the camera and world/region coordinate system (world and region are not rotated related to each other), $(x, y, z)$ is the current camera pose in world coordinate system, $(X_r, Y_r, Z_r)$ are the coordinates of the region's origin in the world coordinate system.

\subsection{Global Mapping}
Following the iMAP work, our approach utilizes the MLP network to implicitly represent a map. The usage of a single MLP has limitations to the working region size due to the limitation in neural network generalization capacity, in other words, in the ``remembering" ability. We propose to use multiple MLP networks that are distributed in the environment. Thus, each single MLP network represents only a part of the environment, e.g. some region, assigned to this network. The local map is the map of such region. The global map can be obtained by merging local maps that are stored in the implicit form of neural network parameters.

The global mapping module consists of two processes. The first one, the networks distribution process, divides the environment into regions, while the second one, the region tracking process, monitors the current pose to define which local map should be involved in the joint optimization process. Each local map has intersections with the neighboring regions. Fig. \ref{preview} demonstrates the global map that consists of green (left) and blue (right) local maps, and their intersection in the middle. The size of each local map is fixed and constitutes 30 m$^3$; the neighboring regions have 20\% of intersections with each other.

When MeSLAM is launched, the MLP network distribution process creates the first local map around the initial pose. During operation, it adds new local maps when the pose enters the intersection area, and initializes the new neural network in the new region using the parameters of the previous local map. For a visited region, MeSLAM uses corresponding network parameters stored for this region.

At each time moment, MeSLAM uses only one MLP for joint optimization, that decreases computational load relative to the use of all intersected MLPs at the same time. The knowledge of which region to use for join optimization is provided by the region tracking process. However, when a pose is in the intersection area, MeSLAM uses the working MLP for joint optimization, and the other one from the intersection is trained but does not influence pose optimization. When a pose crosses the middle of an intersection, the following region becomes the working one and is used in the joint optimization process. 
The region working in joint optimization is depicted in Fig. \ref{sys_overview} in red square.

\subsection{Reconstruction}

The proposed global mapping procedure corrects the shift between local maps. An exponential smoothing and mean prediction in the intersected regions are used. Fig. \ref{reconstr-comp} demonstrates the reconstruction result before and after the smoothing. In the intersection, we use mean predictions with weights exponentially decreased from the region center. The proposed stitching technique creates a smooth and continuous global map, and removes the artifacts caused by different quality of each neural network training.

\section{Experiments and Results}
To validate the performance of the proposed MeSLAM, three sets of experiments were carried out. The first one is the evaluation and comparison of the accuracy of the proposed approach and state-of-art SLAM methods. The second experiment is devoted to the comparison of iMAP and the proposed algorithm performance in a large scale environment. Finally, the third experiment is aimed at the comparison of map sizes obtained by state-of-the-art and proposed MeSLAM approaches.


\subsection{TUM Evaluation}
\begin{table}[!t]
\centering
\caption{ATE RMSE on TUM RGB-D Dataset for State-of-the-art Methods }
\begin{tabular}{llclclc}
\hline
\multicolumn{2}{c}{}                & \multicolumn{2}{c}{fr1/desk (cm)} & \multicolumn{2}{c}{fr2/xyz (cm)} & fr3/office (cm) \\ \hline
\multicolumn{2}{l}{iMAP}            & \multicolumn{2}{c}{4.9}           & \multicolumn{2}{c}{2.0}          & 5.8             \\
\multicolumn{2}{l}{iMAP*}           & \multicolumn{2}{c}{14.8}          & \multicolumn{2}{c}{7.78}         & 11.5            \\
\multicolumn{2}{l}{MeSLAM (ours)} & \multicolumn{2}{c}{6.0}           & \multicolumn{2}{c}{6.54}         & 7.5             \\
\multicolumn{2}{l}{ORB-SLAM2}       & \multicolumn{2}{c}{\textbf{1.6}}  & \multicolumn{2}{c}{\textbf{0.4}} & \textbf{1.0}    \\
\multicolumn{2}{l}{NICE-SLAM}       & \multicolumn{2}{c}{2.7}           & \multicolumn{2}{c}{1.8}          & 3.0             \\
\multicolumn{2}{l}{Kintinuous}      & \multicolumn{2}{c}{3.7}           & \multicolumn{2}{c}{2.9}          & 3.0             \\ \hline
\end{tabular}
\label{tum-evaluation}
\end{table}

TUM RGB-D dataset \cite{sturm2012benchmark} is used to evaluate and compare the accuracy of proposed approach with state-of-the-art SLAM methods: iMAP \cite{sucar2021imap}, iMAP$^*$ (re-implementation of iMAP that is not open-source), ORB-SLAM2 \cite{mur2017orb}, NICE-SLAM \cite{zhu2021nice} and Kintinuous \cite{whelan2012kintinuous}. The calculated metric of SLAM accuracy is Absolute Trajectory Error (ATE RMSE) in cm.

\begin{figure}[!t]
\centerline{\includegraphics[width=8.7 cm]{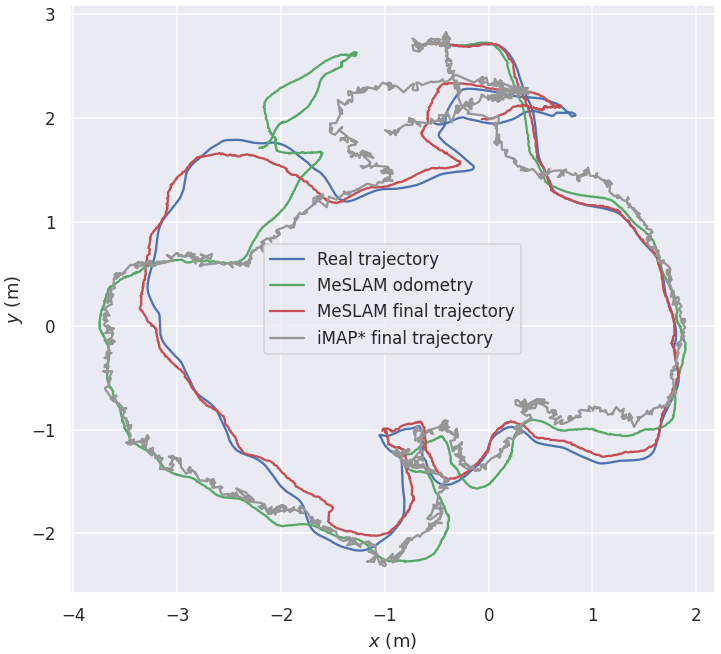}}
\caption{Trajectories of MeSLAM, iMAP$^*$, MeSLAM standalone odometry, and ground truth trajectory on TUM RGB-D dataset, \textit{fr3/office} sequence}
\label{tum-trajectories-comparison}
\end{figure}

The results of the experiment are summarized in Table \ref{tum-evaluation}. In general, MeSLAM shows lower accuracy than state-of-art ORB-SLAM2 metrics, although the values are comparable  (e.g. 6.0 cm against 1.6 cm in \textit{fr1/desk} sequence). In addition, MeSLAM performs more accurately than iMAP* in all sequences and, furthermore, demonstrates stable accuracy results regardless of sequence thanks to the usage of odometry in MeSLAM. The standalone odometry has significant drift within even a short time. However, the combination of external odometry with neural-based optimization provides more accurate, smooth and converged to ground truth trajectories compared to the standalone neural field-based method utilized in iMAP. The example of trajectories is demonstrated in Fig. \ref{tum-trajectories-comparison}.

\subsection{Large Scene Evaluation}\label{exp2}

\begin{figure}[!t]
\centerline{\includegraphics[width=8.7 cm]{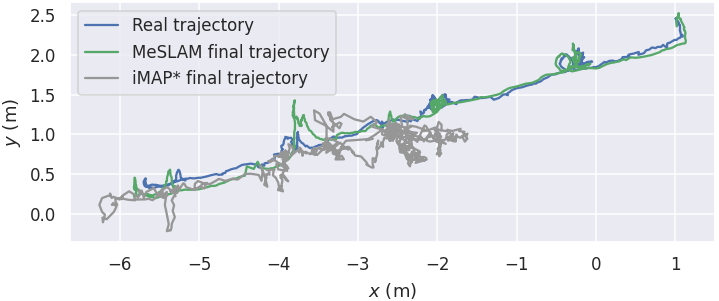}}
\caption{MeSLAM trajectory vs. iMAP in large environment}
\label{trajectories-comparison}
\vspace{-1.5em}
\end{figure}

\begin{figure}[!t]
\centerline{\includegraphics[width=8.7 cm]{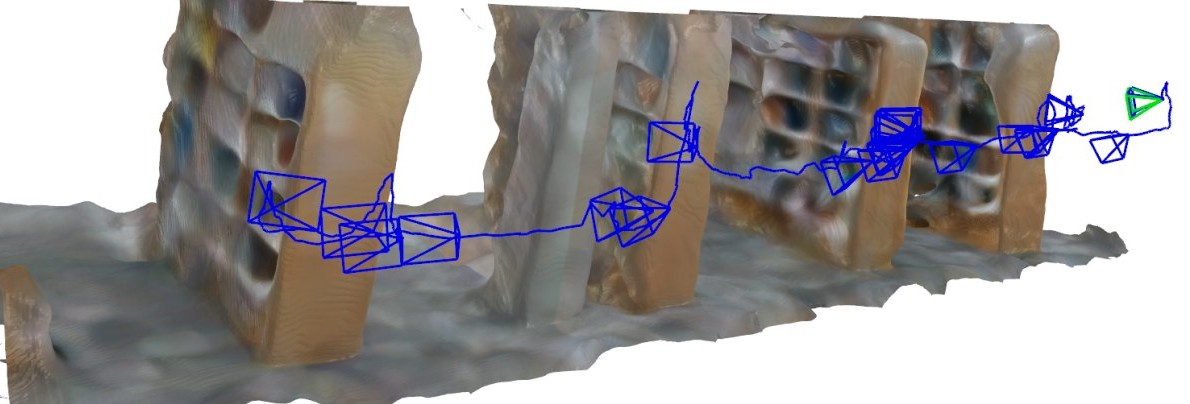}}
\caption{MeSLAM map in a large environment, library}
\label{lib-image}
\vspace{-1em}
\end{figure}

To evaluate and compare the performance of the proposed MeSLAM and state-of-the-art iMAP$^*$ in large environments, an indoor dataset consisting of a long trajectory was recorded using a RealSense D435 camera; both colored and depth data are captured at 15 frames per second with a resolution of 640x480. Both methods are launched on the collected dataset. The estimated and ground truth trajectories are presented in Fig. \ref{trajectories-comparison}. iMAP$^*$ loses its tracking after 60\% of the dataset path and is not able to adequately reflect the map of the environment. Such failure happens because of limitations in the ``remembering" capacity of its neural network. It is impossible to fit large-scale scenes into a small network. Contrariwise, the proposed MeSLAM successfully performs both tracking and mapping due to the networks distribution strategy. The final global map produced by MeSLAM is presented in Fig. \ref{lib-image}.


\subsection{Map Size Evaluation}

\begin{table}[t!]
\caption{Map Size in MB for State-of-the-art Methods}
\begin{center}
\begin{tabular}{lccccc}
\hline
Method                          & \multicolumn{5}{c}{Sequence duration}                                      \\
                                & \textbf{30s} & \textbf{60s} & \textbf{90s} & \textbf{120s} & \textbf{150s} \\ \hline
KinectFusion {[}MB{]}           & 70.1         & 98.9         & 158.2        & 276.7         & 436.6         \\
ElasticFusion {[}MB{]}          & 17.5         & 29.0         & 38.2         & 53.5          & 63.6          \\
RTAB-Map {[}MB{]}               & 8.7          & 17.0         & 25.1         & 33.1          & 37.9          \\
BAD SLAM {[}MB{]}               & 13.1         & 50.0         & 86.4         & 109.1         & 123.8         \\
ORB-SLAM2 {[}MB{]}              & 18.9         & 21.9         & 30.4         & 34.4          & 38.8          \\ \hline
\textbf{MeSLAM (ours) {[}MB{]}} & \multicolumn{5}{c}{\textbf{1.9}}                                           \\ \hline
\end{tabular}
\label{table:map_size}
\end{center}
\end{table}

In order to evaluate the map sizes, we collected a large-scale indoor dataset sequences that consist of images and depth maps using the same setup as in \ref{exp2} experiment. 
 The location captured on each sequence is the same with volume 57 m$^3$. However, the duration of each sequence varies from 30 to 150 sec with a step of 30 sec.
 Thus, a longer sequence can provide more details about the same scenes due to the bigger amount of images and depth maps taken from diverse angles of view, i.e. longer operation time in the volume.

Four dense and one sparse state-of-the-art SLAM methods: KinectFusion \cite{newcombe2011kinectfusion}, ElasticFusion \cite{whelan2015elasticfusion}, RTAB-Map \cite{labbe2019rtab}, BAD-SLAM \cite{schops2019bad} and ORB-SLAM2 \cite{mur2017orb}, were launched on each dataset sequence along with our approach, MeSLAM. The measurable parameter is the size of the obtained 3D map in MB.

\begin{figure}[!t]
\centerline{\includegraphics[width=8.7 cm]{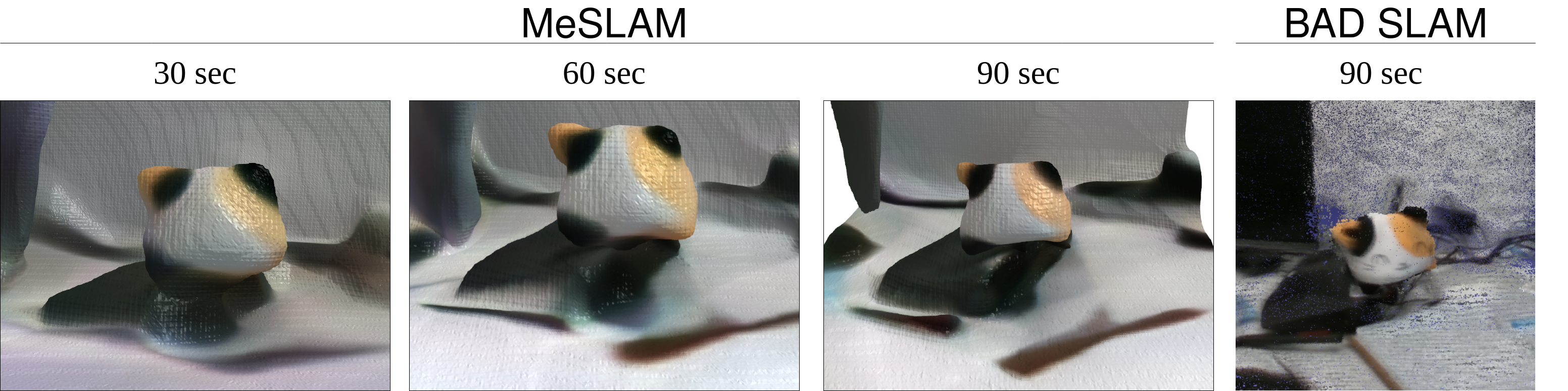}}
\caption{Map representation over time for MeSLAM and BAD SLAM. MeSLAM represents the environment by continuous function and improves the quality over time without increasing required memory, BAD SLAM represents the environment as dense points that increases required memory when the amount of points is increased}
\label{kitties}
\vspace{-1.5em}
\end{figure}

Table \ref{table:map_size} demonstrates the common trend for all methods except the proposed one: the size of the map increases if the operation duration is longer. The proposed method, MeSLAM, shows the fixed size of the map (1.9 MB for 57 m$^3$) because it is stored in the implicit representation as network parameters. Despite the fact that a longer sequence results in better parameters training and, therefore, more detailed map representation depicted in Fig. \ref{kitties}, the map, stored as network parameters, does not grow in its size. However, when a map is stored in explicit representation, e.g. point cloud (all methods except ours), its size grows with the increasing level of scene details; thus, it is costly in terms of disk space. The most efficient map storage among listed state-of-the-art methods is obtained by the RTAB-Map algorithm. Although the proposed approach outperformed the map size of RTAB-Map 4.5 and 19.9 times in 30 sec and 150 sec sequences, MeSLAM maps, being continuous and smooth, are still inferior to the most advanced in terms of the level of map details BAD SLAM method. Fig. \ref{kitties} the comparison of map details for BAD SLAM and MeSLAM approaches. 

\section{Conclusion and Discussion}
In this work, we presented a novel dense neural-based SLAM approach, MeSLAM, for large scale environments with low memory consumption. Firstly, the utilization of odometry alongside the joint optimization process provides precise and stable accuracy of the proposed method. Experimental results have shown that MeSLAM has on average 6.6 cm ATE on TUM RGB-D sequences, which is comparable to the state-of-the-art methods, ORB-SLAM2, NICE-SLAM and Kintinuous, and outperforms the baseline iMAP$^*$ with average accuracy 11.8 cm. Secondly, MeSLAM's global mapping module, that includes the networks distribution and region tracking processes, enables precise mapping of large scenes by decomposing it into regions and stitching them at the time of reconstruction. Experimental results have demonstrated that such an approach surpasses the original iMAP method in the capability of precise tracking and mapping in the large environment. Finally, it is validated that the proposed approach is able to efficiently store the large-sized maps: it requires only 1.9 MB to store 57 m$^3$ while other state-of-the-art methods occupies from 8.7 to 436.6 MB of space to store the same-sized scene.

In future work, authors consider development of bundle adjustment for a global map representation consisting of intersected parts.

\section*{Acknowledgements} 
The reported study was funded by RFBR and CNRS according to the research project No. 21-58-15006.

\bibliographystyle{IEEEtran}
\bibliography{literature}

\end{document}